\documentclass[letterpaper, 10 pt, conference]{ieeeconf}  % Comment this line out if you need a4paper

\IEEEoverridecommandlockouts                              % This command is only needed if 
\usepackage{amsmath}
\usepackage{comment}
\usepackage{amssymb}
\usepackage{array}
\usepackage{here}
\usepackage{enumerate}
\usepackage{bm}
\usepackage{url}
\usepackage{subcaption}
\usepackage{multirow}
\usepackage{mathtools}
\usepackage{url}
\usepackage{graphics} % for pdf, bitmapped graphics files

\title{\LARGE \bf
Integrated Grasping Controller Leveraging Optical Proximity Sensors for Simultaneous Contact, Impact Reduction, and Force Control}

\author{Shunsuke Tokiwa, Hikaru Arita, \emph{Member}, \emph{IEEE}, Yosuke Suzuki, \emph{Member}, \emph{IEEE}, and Kenji Tahara, \emph{Member}, \emph{IEEE}% <-this % stops a space
\thanks{*This work was partially supported by JSPS KAKENHI Grant Number JP20K14702.}% <-this % stops a space
\thanks{S. Tokiwa, H. Arita and K. Tahara are with Department of Mechanical Engineering, Kyushu University, Fukuoka 819-0395, JAPAN (e-mail: tokiwa@hcr.mech.kyushu-u.ac.jp; [arita, tahara]@ieee.org), (fax: (092)802-3225). (Shunsuke Tokiwa and Hikaru Arita are co-first authors.) (Corresponding author: Hikaru Arita.)}
\thanks{Yosuke Suzuki is with the Faculty of Frontier Engineering, Institute of Science and Engineering, Kanazawa University, Kanazawa, Ishikawa 920-1192, JAPAN (e-mail: suzuki@se.kanazawa-u.ac.jp).}}

\begin{document}
\maketitle
\thispagestyle{empty}
\pagestyle{empty}

\begin{abstract}
%%Object grasping is one of the most important tasks for robots.
%%Currently, robots used in factory production lines specialize in grasping known objects.
%%Here, information that is known means the appropriate grasping position and grasping force derived from geometric characteristics such as position and shape, or mechanical characteristics such as softness.
%%On the other hand, the grasping ability of an object whose such information is unknown is not sufficient.
%%Here, we show a method to realize the three functions of simultaneous finger contact, impact reduction, and contact force control, which are effective for grasping an unknown object.

Grasping an unknown object is difficult for robot hands.
When the characteristics of the object are unknown, knowing how to plan the speed at and width to which the fingers are narrowed is difficult.
In this paper, we propose a method to realize the three functions of simultaneous finger contact, impact reduction, and contact force control, which enable effective grasping of an unknown object.
We accomplish this by using a control framework called multiple impedance control, which was proposed in a previous study.
The advantage of this control is that multiple functions can be realized without switching control laws.
The previous study achieved two functions, impact reduction and contact force control, with a two layers of impedance control which was applied independently to individual fingers.
In this paper, a new idea of virtual dynamics that treats multiple fingers comprehensively is introduced, which enables the function of simultaneous contact without compromising the other two functions.
This research provides a method to achieve delicate grasping by using proximity sensors.
For the effectiveness of the proposed method, please refer to \url{https://youtu.be/q0OrJBal4yA}.

\end{abstract}
\begin{keywords}
Force control, sensor-based control, grasping, optical proximity sensor.
\end{keywords}

%%%%%%%%%%%%%%%%%%%%%%%%%%%%%%%%%%%%%%%%%%%%%%%%%%%%%%%%%%%%%%%%%%%%%%%%%%%%%%%%
\section{INTRODUCTION}
\subsection{Motivation}
\noindent
\PARstart{T}{HE} grippers currently in use on factory production lines often specialize in grasping objects with known geometric characteristics, such as position and shape, and mechanical characteristics, such as softness.
For example, industrial products are made according to design data, so their shape, size, and composition are almost constant.
In addition, since many of these products are relatively hard, large grasping forces rarely destroy or damage them.
In contrast, food and agricultural products vary in shape, size, and composition from object to object.
Many of them are relatively soft, and application of a large grasping force can lead to their destruction or damage.
In addition, there are cases in which the product rolls and changes position because it is round and cases in which the surface is membranous and does not permit rubbing.
These features make grasping difficult\cite{hayashi}, \cite{wang}.
Note that the above problems are not limited to food and agricultural products but can also occur with irregularly shaped bagged products.

%%If the characteristics of the object are constant, it is possible to deal with the object by repeating the same operation, such as moving the gripper to the same position, closing the gripper fingers at the same speed, and grasping the object with the same force.
%%If the characteristics of the object are different each time, but the characteristics can be known accurately in advance, it is possible to deal with the differences in the characteristics of the object by using multiple grippers or by changing the width and speed of the gripper closing each time.
%%On the other hand, the capacity to handle cases in which the characteristics of the object are different each time and cannot be accurately known in advance is not sufficient.
%%An example is the grasping of food and agricultural products.
%%Unlike machine parts, such as screws and gears, which have a fixed size, shape, and material, food and agricultural products have individual differences even in the same type.
%%Under these conditions, even if multiple grippers are available, it is not possible to select the appropriate gripper, and it is also difficult to pre-set the width and speed of gripper closing for each object.
\par
We believe that the above problem could be effectively solved by providing the following three functions in the grasping process.
\begin{itemize}
\item Simultaneous contact of each finger with the object
\item Reduction of the impact when the fingers contact the object
\item Control of the contact forces acting on the object
\end{itemize}
%%The reason why these functions are effective is considered using food as an example of an object that is difficult to handle with the aforementioned current grasping capacity.
%%First, the simultaneous contact function enables grasping of food without shifting, thereby reducing the risk of food toppling or food damage due to undesirable forces such as frictional forces generated between the food and the floor surface.
%%For example, when grasping a cod roe whose skin is easily torn, the skin may be damaged by frictional force and the contents may come out.
%%To prevent this, it is important to have a simultaneous contact function that enables grasping without shifting the object's position.
%%The impact reduction function also reduces the risk of destruction or damage to the food due to the impact force when the gripper fingers make contact with the foods.
%%For example, a fragile object such as tempura batter is easily damaged by even the slightest force, and it is important to grasp it without generating a large force, even momentarily.
The reasons why we consider these functions to be effective are as follows.
First, the simultaneous contact function enables grasping without shifting, thereby reducing the risk of toppling or damage due to undesirable forces such as frictional forces generated between the object and the floor surface.
Second, the impact reduction function reduces the risk of destruction of or damage to the object due to the impact force that occurs when the gripper fingers make contact with the object.
Furthermore, force control-based grasping after contact enables excessive contact forces on the object to be avoided even under conditions in which the exact size and shape of the object, such as food, is unknown.
\par
Each of the three functions listed above can be achieved by existing methods (details are explained in Sec.\,\rm{I}.\textit{B}), but there is no method that integratively achieves them.
The reason why the three functions cannot be integratively realized by combining existing methods is that the base control method (position/force) is different for each function.
This requires switching of control laws, which causes system instability and increased contact time.
%Editor: Please ensure that the intended meaning has been maintained in the above edit.OK
In this study, we utilize the control framework proposed by Arita et al.\cite{arita}, i.e., ``multiple impedance control,'' to integrate the three functions.
The concept of a virtual force is introduced in multiple impedance control, enabling the recognition of sensor information other than that from force sensors as force information.
The control method\cite{arita} recognizes the output of a proximity sensor as force information and achieves impact reduction.
Furthermore, by connecting the impedance control in series, contact force control is smoothly transitioned to after the control target contacts the object.
Thus, in multiple impedance control, each function is realized without switching control laws by realizing the multiple functions on a force-control basis.
\par
In this study, the three functions are realized by adding simultaneous contact control to the impact reduction and contact force control\cite{arita}.
The key idea is the introduction of a virtual dynamics that treats all fingers comprehensively. Needless to say, simultaneous contact is not something that can be achieved by individual fingers alone, but rather its success or failure is determined by the interaction of all fingers. This differs from the impact reduction and contact force control, which can be accomplished by individual fingers. Therefore, in this study, we consider a more abstract virtual dynamics and virtual force, and achieve our objective by introducing a virtual dynamics with respect to the geometric center coordinates of all fingers.
Furthermore, the simultaneous contact control added in this study is realized without compromising the impact reduction and contact force control function\cite{arita}.
This fact shows that each function can be designed independently and realized in an integrated manner by considering the virtual dynamics corresponding to each function.

\subsection{Related works}
Various grasp planning methods have been proposed, and one example is a method that uses a vision sensor\cite{vision1}-\cite{vision6}.
However, vision sensors have the disadvantage that necessary information may be missing due to occlusion, and even if occlusion can be avoided, there is a delay in accessing specific information due to the abundance of information available from vision sensors.
Such delays and missing information make achieving simultaneous contact and impact reduction with vision sensors difficult.
Therefore, intelligent hand and grasp planning methods using proximity sensors that enable information about the robot vicinity to be obtained have been proposed\cite{search_grasp_prox}-\cite{koyama_impact}.
Among them, the method proposed by Koyama et al.\cite{koyama_simultaneous}, \cite{koyama_impact} is highly relevant to this study.
In \cite{koyama_simultaneous}, simultaneous contact of each finger is achieved by using an optical proximity sensor.
The output of an optical proximity sensor varies depending on the reflectance of the object, which is independent of simultaneous contact.
Therefore, the reflectance is estimated from the change in the sensor output while closing the fingers, the position of each finger is controlled at an equal distance from the surface of the object by using the sensor output that is eliminated reflectance effects, and simultaneous contact is achieved.
In \cite{koyama_impact}, the distance to the object is detected by a customized proximity sensor, and the derivative of the detected distance is defined as the virtual damping force.
Impact reduction is achieved by applying the virtual damping force to each finger of the robot hand.
This approach has succeeded in grasping soft objects such as marshmallows and paper balloons at high speeds with only minute deformation.
\par
Although simultaneous contact and impact reduction functions have been realized, these functions have not been realized with a single control law.
Additionally, these control approaches are position control methods in which the relative position between the fingertip and the object is controlled, and no method with a smooth transition to force control after achieving these functions existed before \cite{arita} proposed one.
Therefore, switching of control laws is required to realize the three functions of simultaneous contact, impact reduction, and contact force control using existing methods.
The multiple impedance control used in this study eliminates this problem and is an extension of the series admittance-impedance control proposed by Fujiki et al\cite{fujiki}.
In series admittance-impedance control, only the contact force is used as an input, but in multiple impedance control, the control structure is generalized by using the concept of a virtual force.
In this study, the three functions are realized by using multiple impedance control for the first time.

\section{Overview of the control target and proximity sensors}
In this study, a two-finger parallel gripper is used as the control target as the simplest example.
Optical proximity sensors are used to realize simultaneous contact and impact reduction.
The proximity sensors in this study are reflected light intensity-type (O-RLI-type) sensors composed of photoreflectors.
A sensor is mounted on each finger of the gripper to acquire information about the object and reflect it in the control law.
Each finger of the gripper can be treated as an independent 1-degree-of-freedom (DoF) system.
A gripper in which each finger can be independently controlled was proposed by Sato et al\cite{sato}.
The method proposed in this letter can be directly applied to such a gripper.
Even if the gripper is a general gripper in which every finger moves in tandem, equivalent control is possible by controlling the position of the wrist of the arm.
Considering that the $x$ coordinate is attached to the gripper wrist, the motion of each finger is modeled as follows:
\begin{equation}
\label{motion_equ}
	m_j\ddot{x}_j=u_j+f_{\text{c}j}.
\end{equation}
where $m_j$ is the mass of the gripper finger, $x_j$ is the fingertip position, $u_j$ is the control input, $f_{\text{c}j}$ is the contact force acting from the grasped object, and subscripts $j=1$ and $2$ indicate the right and left fingers of the gripper, respectively.
%%The control method described below achieves simultaneous contact, impact reduction, and contact force control during grasping by determining the control input $u_j$ in (\ref{motion_equ}).

\section{Proposed control method}
\subsection{Simultaneous contact, impact reduction, and contact force control} 
\begin{figure*}[htbp]
\centering
\includegraphics[width=150mm]{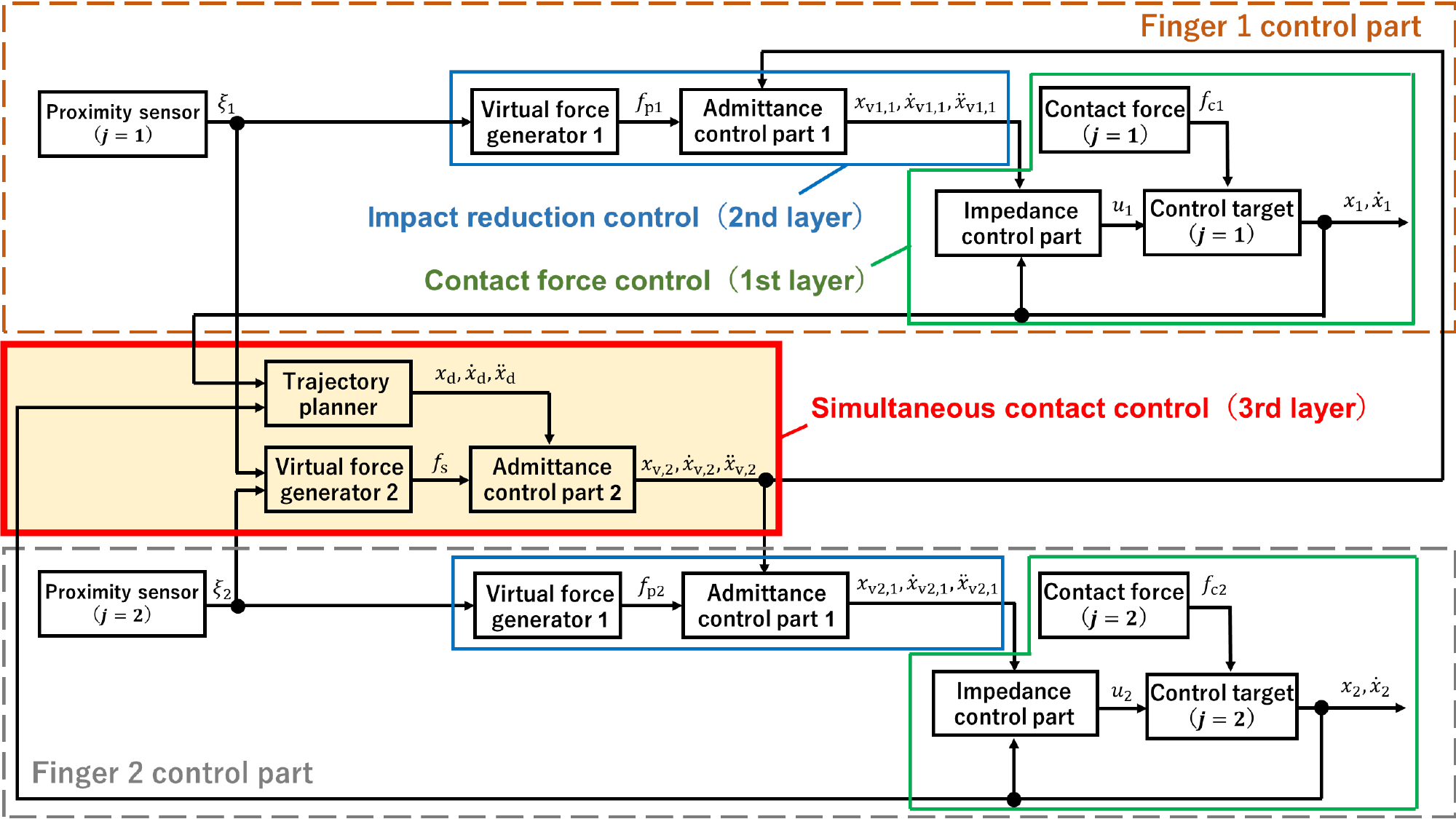}
\caption{Block diagram of the proposed controller using multiple impedance control. It consists of three layers of impedance control, with each layer playing the role of simultaneous contact, impact reduction, or contact force control.}
\label{proposal_method}
\end{figure*}
A block diagram of the control law proposed in this study is shown in Fig.\,\ref{proposal_method}.
Each finger of the gripper is controlled by multiple impedance control, which consist of three layers.
From the top layer, these layers play the roles of simultaneous contact, impact reduction, and contact force control.
The bottom two layers (impact reduction and contact force control) are the same as those in the control method proposed in \cite{arita}, and in this study, one control layer for simultaneous contact is added to the top.
The third layer is an admittance control layer, which plays the role of simultaneous contact control.
The virtual force $f_\text{s}$ calculated by virtual force generator 2 is given by
\begin{equation}
	f_\text{s}=K_\text{s}\text{tanh}\biggl(\frac{\xi_1-\xi_2}{\xi_1+\xi_2}\biggr),
\label{fs}
\end{equation}
where $\xi_j$ is the output of the proximity sensor and $K_\text{s}$ is the gain.
Note that since an O-RLI-type proximity sensor is used, the output $\xi_j$ increases in an analog-manner when approaching an object.
$f_\text{s}$ is used in admittance control part 2 as an external force on the virtual object and is determined by the following equation of motion:
\begin{equation}
	M_{\text{a},2}(\ddot{x}_{\text{v},2}-\ddot{x}_\text{d})+D_{\text{a},2}(\dot{x}_{\text{v},2}-\dot{x}_\text{d})+K_{\text{a},2}(x_{\text{v},2}-x_\text{d})=f_{\text{s}},
\label{proposal_equ}
\end{equation}
where $M_{\text{a},2}$, $D_{\text{a},2}$, and $K_{\text{a},2}$ are the mass, viscosity coefficient, and elastic coefficient of the virtual object, and $x_{\text{v},2}$, $\dot{x}_{\text{v},2}$, and $\ddot{x}_{\text{v},2}$ are the position, velocity, and acceleration of the virtual object determined by admittance control part 2.
Additionally, $x_{\text{d}}$, $\dot{x}_{\text{d}}$, and $\ddot{x}_{\text{d}}$ are the target trajectories, which are given by the trajectory planner as follows:
\begin{equation}
	x_\text{d}=\frac{x_1+x_2}{2},\ \dot{x}_\text{d}=0,\ \ddot{x}_\text{d}=0,
\label{xd}
\end{equation}
where $x_\text{d}$ represents the geometric center of gravity of the gripper.
The idea of using the geometric center of gravity of the gripper as the target trajectory is based on the method proposed by Ozawa et al.\cite{grasp} for stable grasping without requiring information on the object.
In other words, the admittance control in the third layer can be considered to play the role of correcting the target trajectory for simultaneous contact using the output of the proximity sensor.
\par
The second layer is an admittance control layer that plays the role of impact reduction.
The second layer uses the position $x_{\text{v},2}$, velocity $\dot{x}_{\text{v},2}$ and acceleration $\ddot{x}_{\text{v},2}$ of the virtual object obtained in the third layer as the desired state.
The virtual viscous force $f_{\text{p}j}$ calculated by virtual force generator 1 is given by
\begin{equation}
	f_{\text{p}j}=D_{\text{p}j}\frac{\dot{\xi}_j}{\xi_j},
\label{fpj}
\end{equation}
where $D_{\text{p}j}$ is the viscosity coefficient.
$f_{\text{p}j}$ is used in admittance control part 1 as an external force on the virtual object and is determined by the following equation of motion:
\begin{equation}
	M_{\text{a},1}(\ddot{x}_{\text{v}j,1}-\ddot{x}_{\text{v},2})+D_{\text{a},1}(\dot{x}_{\text{v}j,1}-\dot{x}_{\text{v},2})+K_{\text{a},1}(x_{\text{v}j,1}-x_{\text{v},2})=f_{\text{p}j},
\label{impact_reduction_ad}
\end{equation}
where $M_{\text{a},1}$, $D_{\text{a},1}$, and $K_{\text{a},1}$ are the mass, viscosity coefficient, and elastic coefficient of the virtual object and $x_{\text{v}j,1}$, $\dot{x}_{\text{v}j,1}$, and $\ddot{x}_{\text{v}j,1}$ are the position, velocity, and acceleration of the virtual object determined by admittance control part 1.
\par
The first layer is an impedance control layer that plays the role of contact force control.
The first layer uses the position $x_{\text{v}j,1}$ and velocity $\dot{x}_{\text{v}j,1}$ of the virtual object obtained in the second layer as the desired state.
Since no force sensor is mounted on the fingers in this research, no inertia-related control is performed, and $m_j$ is left as is.
In this case, $u_j$ is given as follows:
\begin{equation}
\label{input}
	u_j=-D_\text{i}(\dot{x}_j-\dot{x}_{\text{v}j,1})-K_\text{i}(x_j-x_{\text{v}j,1}),
\end{equation}
where $D_\text{i}$ and $K_\text{i}$ are the desired viscosity and elastic coefficients in impedance control.
Therefore, $u_j$ as expressed in (\ref{input}) is used to rewrite the equation of motion of the control target (\ref{motion_equ}) as follows:
\begin{equation}
	m_j\ddot{x}_j+D_\text{i}(\dot{x}_j-\dot{x}_{\text{v}j,1})+K_\text{i}(x_j-x_{\text{v}j,1})=f_{\text{c}j}.
\label{dn}
\end{equation}
The contact force $f_{\text{c}j}$ acting from the object follows (\ref{dn}).

\subsection{Effect of reflectance on simultaneous contact control}
\begin{table*}[htp]
 \caption{Control parameters}
 \centering
    \scalebox{1.05}{
     \renewcommand{\arraystretch}{1.2}
    \begin{tabular}{cccccccccc}
        \hline
        \multicolumn{4}{c}{Simultaneous Contact Control} & \multicolumn{4}{|c|}{Impact Reduction Control} & \multicolumn{2}{c}{Contact Force Control} \\
        \hline
$M_{\text{a},2}$ & $D_{\text{a},2}$ & $K_{\text{a},2}$ & \multicolumn{1}{c|}{$K_\text{s}$} & $M_{\text{a},1}$ & $D_{\text{a},1}$ & $K_{\text{a},1}$ & \multicolumn{1}{c|}{$D_{\text{p}1,2}$} & $D_{\text{i}}$ & $K_{\text{i}}$ \\
\hline
1 kg & 12 N$\cdot$s/m &  36 N/m & \multicolumn{1}{c|}{3 N} & 1 kg & 10 N$\cdot$s/m & 25 N/m & \multicolumn{1}{c|}{$\pm$0.6 N$\cdot$s} & 14 N$\cdot$s/m & 98 N/m \\
 \hline
\label{parameter}
  \end{tabular}
  }
\end{table*}

\begin{table}[htp]
 \caption{Physical quantities related to the control target and the grasped object}
 \centering
  \scalebox{1.05}{
    \renewcommand{\arraystretch}{1.2}
  \begin{tabular}{ccccc}
   \hline
   $x_{\text{init},1}$ & $x_{\text{init},2}$ & $m_j $ & $W$ & $x_\text{m}$  \\
   \hline
   0.11 m & $-$0.11 m & 0.5 kg & 0.07 m & 0.03 m  \\
  \hline
\label{Initial conditions and physical quantitys}
  \end{tabular}
  }
\end{table}

\begin{figure}[tp]
\centering
\includegraphics[width=80mm]{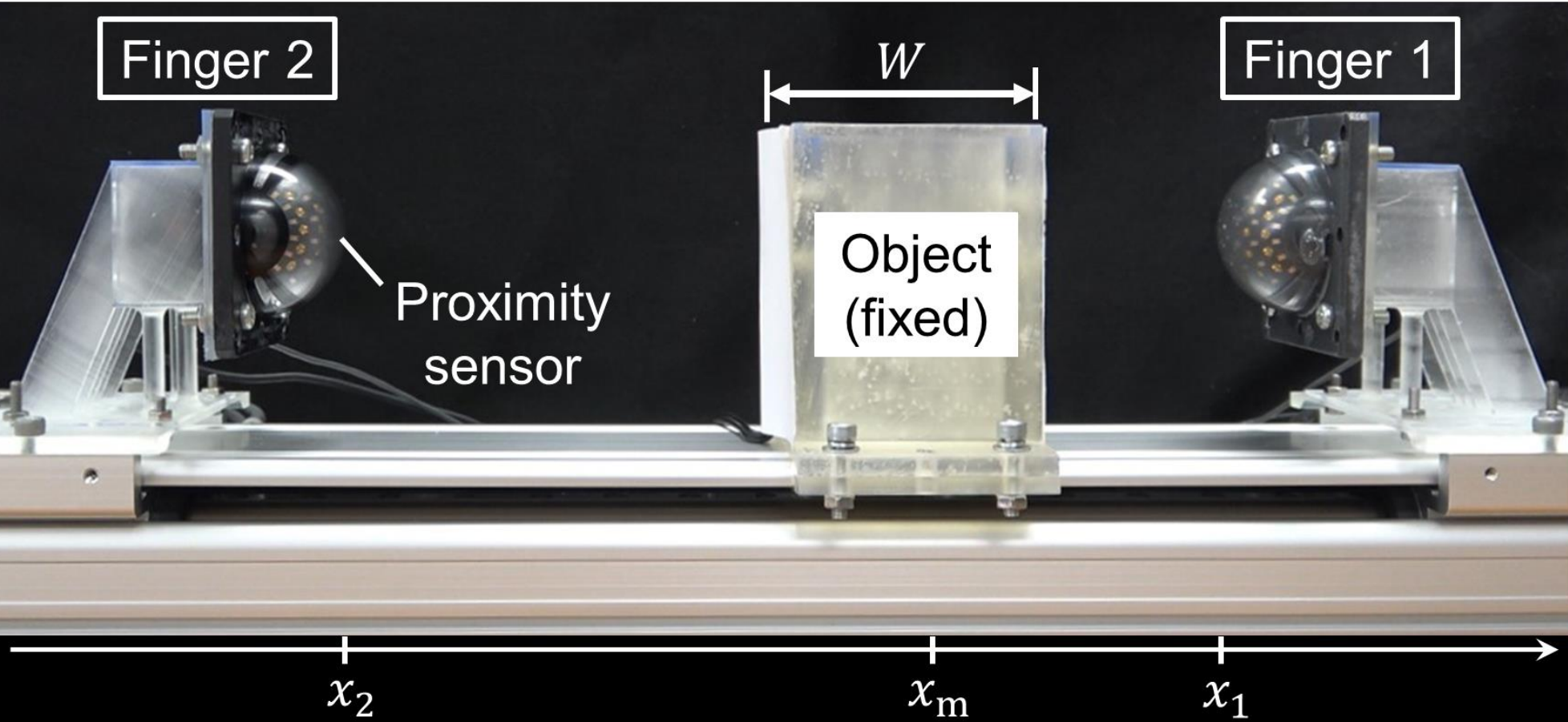}
\caption{Experimental device used to verify whether the proposed method can achieve both the simultaneous contact and impact reduction functions.}
\label{Experimental setup}
\end{figure}

The O-RLI-type sensor used in the proposed method is smaller and has a faster response time than other types of sensors, but its output depends on the reflectance of the object\cite{sensor}.
This section describes the effect of the reflectance of the object surface on simultaneous contact control.
The output of an O-RLI-type sensor is modeled as follows\cite{koyama_model}:
\begin{equation}
	\xi_j=G_{\xi}\frac{\alpha_j\psi}{(l_j+l_0)^n},
\label{sensor1}
\end{equation}
where $G_{\xi}$ is the coefficient for the transformation from the amount of light received to a voltage determined by a sensing element such as a phototransistor, $\psi$ is the light energy emitted from the LED, $l_j$ is the distance between the proximity sensor and the object surface, $l_0$ is the offset distance to prevent direct contact between the sensor element and the object, and $n$ is the diffusion coefficient, which depends on the number and arrangement of LEDs and phototransistors.
Additionally, $\alpha_j$ is the reflectance of the object surface, and the subscript $j$ specifies the object surface to be detected by the proximity sensor of the right or left finger.
Substitution of (\ref{sensor1}) into (\ref{fs}) leads to the following equation:
\begin{equation}
    f_\text{s}=K_\text{s}\text{tanh}\biggl\{\frac{\alpha_1(l_2+l_0)^n-\alpha_2(l_1+l_0)^n}{\alpha_1(l_2+l_0)^n+\alpha_2(l_1+l_0)^n}\biggr\}.
\label{fsa}
\end{equation}
When the reflectances of the object surface detected by the right and left proximity sensors are equal ($\alpha_1=\alpha_2$), the following equation is obtained:
\begin{equation}
	f_\text{s}=K_\text{s}\text{tanh}\biggl\{\frac{(l_2+l_0)^n-(l_1+l_0)^n}{(l_2+l_0)^n+(l_1+l_0)^n}\biggr\}.
\label{fsae}
\end{equation}
According to (\ref{fsae}), the virtual force $f_\text{s}$ is independent of the reflectance $\alpha_j$ if $\alpha_1=\alpha_2$ holds.
In other words, if the reflectances of the object detected by the proximity sensors attached to each finger are equal, then the control law is independent of the reflectance.
Thus, some of the disadvantages of the O-RLI-type sensor, for which the output depends on the reflectance of the object, can be eliminated.

\section{Evaluation}
In this section, the proposed method is evaluated via simulations and experiments.
In the simulations, the evaluation is conducted under ideal conditions, thus eliminating factors that cannot be ignored in the experiment, such as frictional forces acting on the control target.
Additionally, since simulation requires modeling of sensor outputs and contact handling, we also evaluate the proposed method via experiments that do not include such approximate elements.
\par
Fig.\,\ref{Experimental setup} shows an overview of the experimental setup. 
The control parameters and physical quantities related to the control target and the grasped object in the simulation and experiment are shown in Table\,\ref{parameter} and \ref{Initial conditions and physical quantitys}, respectively.
Note that $x_{\text{init},j}$ is the initial position of each finger of the gripper, $W$ is the width of the object, and $x_\text{m}$ is the center position of the object.

\subsection{Simulations}
In this section, the effect of adding simultaneous contact control is evaluated via simulation.
For this purpose, the following two methods are compared and evaluated.
\vskip.5\baselineskip
\paragraph*{Case\,1} \ 
\par
The method proposed in \cite{arita} is applied to each finger of the gripper.
Note that $x_{\text{v},2}$, $\dot{x}_{\text{v},2}$, and $\dot{x}_{\text{v},2}$ in (\ref{impact_reduction_ad}) are replaced by $x_{\text{d}}$, $\dot{x}_{\text{d}}$, and $\ddot {x}_{\text{d}}$ in (\ref{xd}), respectively.
\par
\paragraph*{Case\,2} \ 
\par
The proposed method described in Sec.\,\rm{I}\hspace{-1pt}\rm{I}\hspace{-1pt}\rm{I}.\textit{A} is applied to each finger of the gripper.
\vskip.5\baselineskip
This simulation addresses the contact between the gripper fingers and the object.
Therefore, the equations of motion, including the constraint conditions, must be numerically analyzed, and J. Baumgarte's constraint stabilization method (CSM) is used as the stabilization method\cite{CSM}.
%%To set the color of the object surface to white and the offset distance to $l_0=0.005\ \text{m}$,
%%The color of the object surface is set to white, the offset distance is set to $l_0=0.005\ \text{m}$, and the experimental data are fitted with (\ref{sensor1}).

\subsubsection{Modeling of the proximity sensor output}
\begin{figure}[tp]
\centering
\includegraphics[width=80mm]{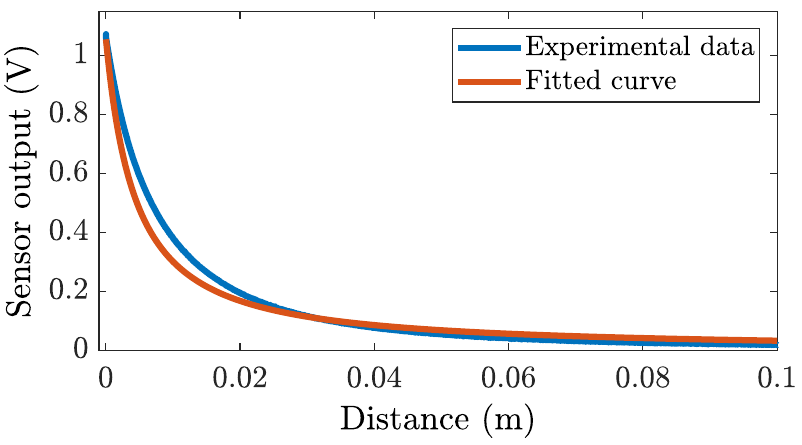}
\caption{Relationship between the sensor--object distance and proximity sensor output when the object surface is white. The blue curve represents the experimental data, and the red curve represents the fitted curve.}
\label{fitted_curve}
\end{figure}

In this simulation, (\ref{sensor1}) is used to model the output of the proximity sensor.
Let $K\coloneqq G_\xi\alpha_j\psi$; then, the fitting parameters required to represent the proximity sensor output are $K$ and $n$.
Also, $l_0$ is the offset distance determined by the sensor design, and is set to $l_0=0.005\ \text{m}$.
From (\ref{sensor1}), the sensor output must be zero when the sensor and the object are sufficiently far apart.
In the fitting and the experiments described in Sec.\,\rm{I}\hspace{-1pt}\rm{V}.\textit{B}, the sensor output is offset to satisfy the above conditions.
The experimental data are fitted with (\ref{sensor1}) for the case where the surface color of the object is white.
As a result, $K=0.0024$, $n=1.1506$, $c=0.35$, and coefficient of determination $R^2=0.9992$ are obtained.
The experimental data and the fitting curve of (\ref{sensor1}) are shown in Fig.\,\ref{fitted_curve}.
According to Fig.\,\ref{fitted_curve}, (\ref{sensor1}) explains the experimental data well.
\begin{figure*}[htbp]
	\centering
	\begin{minipage}[b]{0.45\linewidth}
		\centering
		\includegraphics[width=77mm]{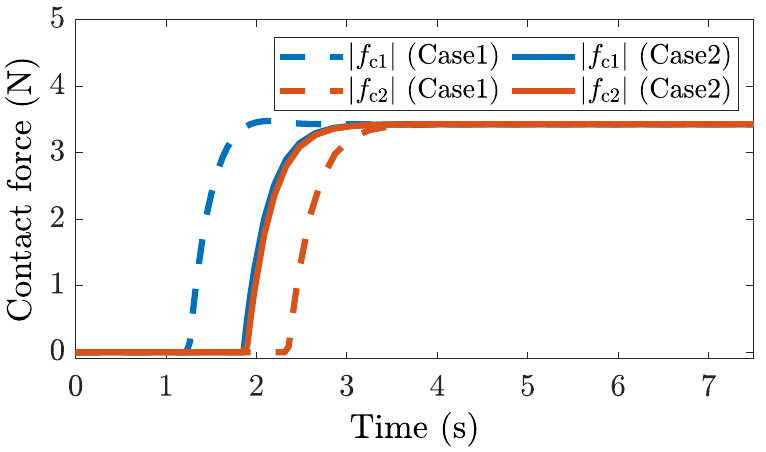}
		\subcaption{Time series data of the contact forces acting on each finger. The blue line shows the contact force acting on finger $j=1$, and the red line shows the contact force acting on finger $j=2$. The dashed line represents the results for Case\,1, and the solid line represents Case\,2.}
	\end{minipage}
         \hspace{0.04\columnwidth}
	\begin{minipage}[b]{0.45\linewidth}
		\centering
		\includegraphics[width=80mm]{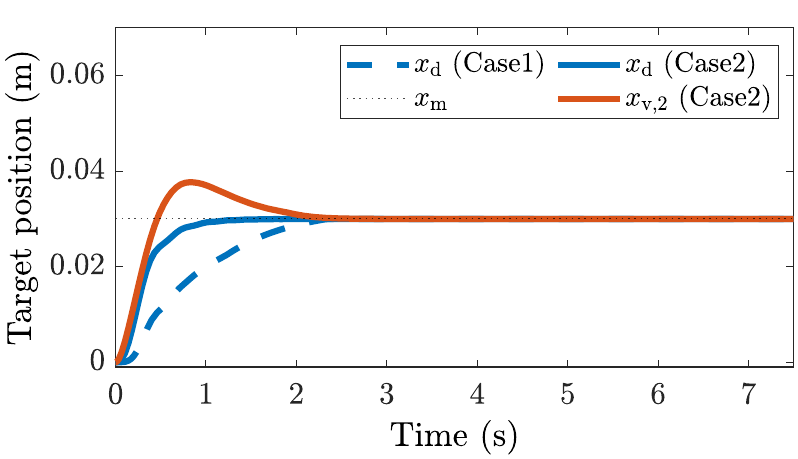}
		\subcaption{Time series data of the target position. The dotted line represents the center position of the grasped object, the dashed line represents the results for Case\,1, and the solid line represents Case\,2.\\ \ }
	\end{minipage}
	\caption{Simulation results with and without simultaneous contact control.}
	\label{simu_comb}
\end{figure*}
\subsubsection{Simulation conditions}
In this simulation, the object surface detected by the proximity sensor is considered to be a plane, the reflectance is assumed to be constant across the entire plane, and the reflectances of the surface detected by the right and left proximity sensors are equal ($\alpha_1=\alpha_2$).
The object is assumed to be rigid, and the position of the object is assumed to be fixed to facilitate evaluation of the simultaneous contact function.
Furthermore, the area of the object surface is assumed to be sufficiently large compared to the detection range of the proximity sensor.

\subsubsection{Simulation Results}
%%The contact force and target position results for Case 1 are shown in Fig. \ref{arita_result_simu2} and the results for Case 2 are shown in Fig. \ref{proposal_result_simu2}.
The contact force and target position results for Cases\,1 and 2 are shown in Fig.\,\ref{simu_comb}.
First, we focus on the time each finger takes to contact the object and the time difference of contact.
In addition to the results obtained in Fig.\,\ref{simu_comb}(a), the simulation results for $x_\text{m}=0.00, 0.01, 0.02, 0.04, 0.05\,\text{m}$ are also summarized in Table\,\ref{position_dep} to evaluate the object position dependence of the simultaneous contact function.
Table\,\ref{position_dep} shows that the time required for finger\,2 (the finger with a longer distance to the object surface) to make contact becomes shorter while the time required for finger\,1 (the finger with a shorter distance to the object surface) to make contact becomes longer when simultaneous contact control is applied.
In other words, simultaneous contact control plays a role in balancing the times each finger takes to make contact.
This effect reduces the time difference of contact and is independent of the center position of the object $x_\text{m}$.
%%This effect reduces the time difference of contact.
%%In addition, Table \ref{position_dep2} shows that when simultaneous contact control is not applied, the time required for Finger 1 to make contact becomes shorter and the time required for Finger 2 to make contact becomes longer as $x_\text{m}$ becomes larger.
%%This increases the time difference of contact.
%%On the other hand, when the simultaneous contact control is added, the time required for both Finger 1 and Finger 2 to make contact does not change significantly, and the contact time difference is reduced.
%%Therefore, the simultaneous contact control can perform its function regardless of the position of the object.
\par
Next, we focus on the target position.
From Fig.\,\ref{simu_comb}(b), we can confirm overshoot of the target position $x_{\text{v},2}$ from the center position of the object when adding simultaneous contact control.
The role of target position overshoot is to move the target position in the direction of the finger with the shorter distance to the object surface so that this finger has less force applied to it to close the finger and the finger with the longer distance to the object has more force applied to it to close the finger.
This balances the times each finger takes to make contact and reduces the time difference of contact, as shown in Table \ref{position_dep}.
The target position $x_{\text{v},2}$ can be confirmed to finally become $x_{\text{v},2}=x_\text{m}=0.03\ \text{m}$ and to converge to the center position of the object.
In other words, the final state is equivalent to that for the method in \cite{grasp}.
\par
These results show that under ideal conditions, in which factors such as friction are eliminated, simultaneous contact control has the effect of balancing the times each finger takes to makes contact.
This reduces the time difference of contact, and we can also confirm that this effect is independent of the position of the object.

\subsection{Experiments}
\begin{figure}[htbp]
\centering
\includegraphics[width=50mm]{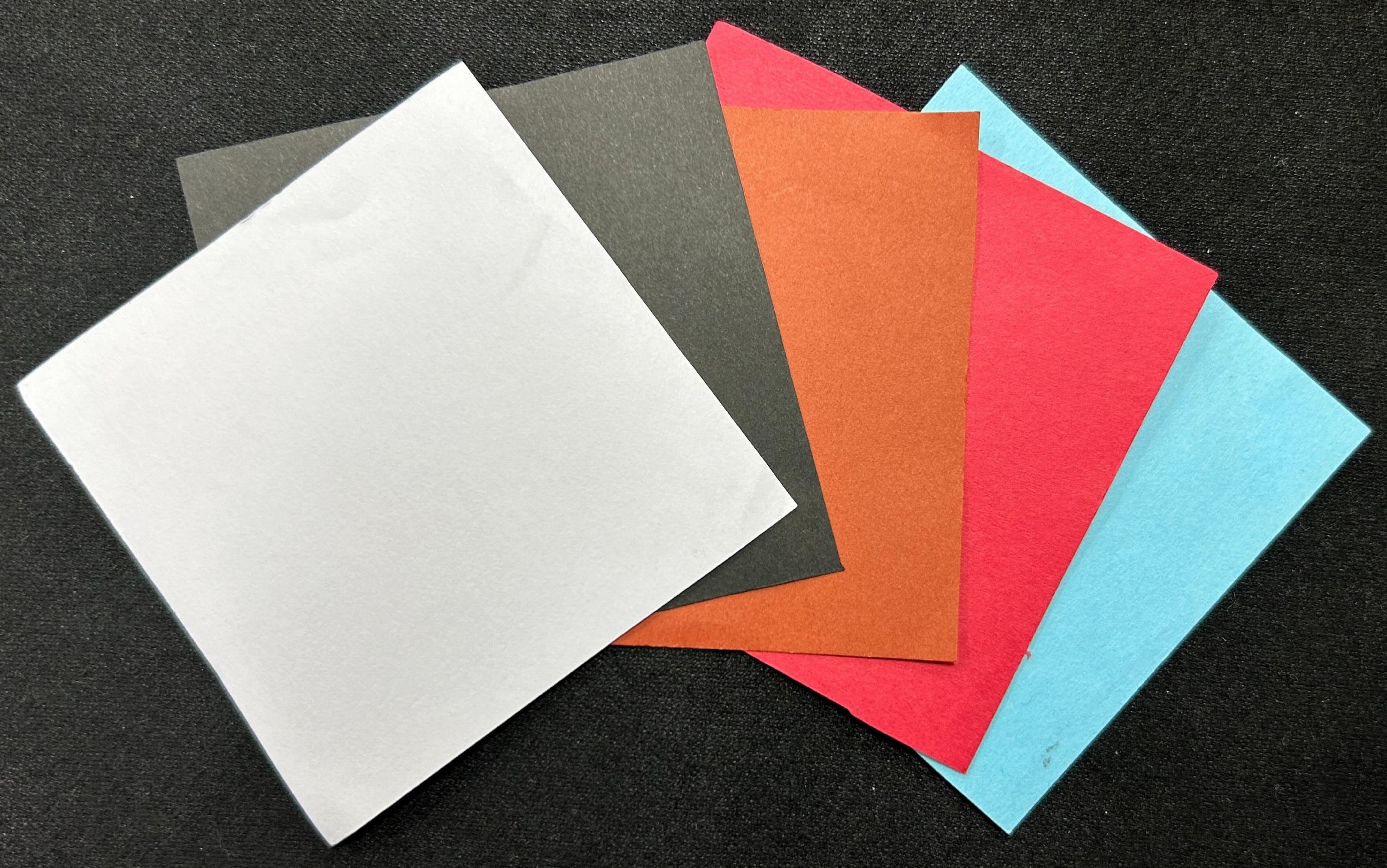}
\caption{Paper of five colors to be attached to the surface of the object. From left to right: white, black, brown, red and blue.}
\label{copa}
\end{figure}
Through experiments on actual equipment, we verify whether the proposed method can achieve the simultaneous contact function without losing the impact reduction function.
For this purpose, in addition to Cases\,1 and 2 described in Sec.\,\rm{I}\hspace{-1pt}\rm{V}.\textit{A}, we also examine Case\,0, which is described below.
\vskip.5\baselineskip
\paragraph*{Case\,0}\ 
\par
The impedance control shown in (\ref{input}) is applied to each finger of the gripper.
Note that $x_{\text{v}j,1}$ and $\dot{x}_{\text{v}j,1}$ are replaced by $x_{\text{d}}$ and $\dot{x}_{\text{d}}$ in (\ref{xd}), respectively.
\vskip.5\baselineskip
%%By comparing the results in Case 0, 1, and 2, we evaluate the success or failure of the proposed method in achieving both simultaneous contact and impact reduction functions.

\subsubsection{Experimental setup}
%%Fig. \ref{Experimental setup} shows an overview of the experimental setup. 
The experimental setup (Fig. 2) includes two O-RLI-type sensors (Create sensors similar to the sensor proposed in \cite{suzuki_grasp}), an object, two force sensors (USL06-H5-200N-C, Tec Gi Han Co., Ltd., Ibaraki, Japan), a dual linear stage (SLP-15-300-D-M3-A3-SH, Nippon Pulse Motor Co., Ltd., Tokyo, Japan), and a real-time controller (MicroLabBox, dSPACE GmbH, Paderborn, Germany).
The dual linear stage can individually move two tables. 
Each table is equipped with a proximity sensor, and each table is considered a finger of a gripper.
The object is fixed on the linear stage, and a force sensor is attached to each surface where the gripper fingers make contact.
The force sensors measure the impact force and time difference of contact to evaluate the impact reduction and simultaneous contact functions.
Additionally, by attaching colored paper to the surface of the object, the reflectance of the surface is changed.
The real-time controller measures the signals of the proximity sensors, force sensors, and encoders mounted on the dual linear stage and controls each table force via a motor driver (MADLT11SM, Panasonic Industry Company, Ltd., Osaka, Japan).
In this experiment, the sampling rate of the controller is set to 10 kHz, and a fifth-order Butterworth low-pass filter with a cutoff frequency of 500 Hz and a median filter with a sample size of 10 are used to reduce the noise of the proximity sensors.

\subsubsection{Experimental conditions for evaluating the reflectance dependence}
To investigate the dependence on the reflectance of the object, experiments are conducted by changing the color of the paper attached to the surface.
The colors of the paper to be attached are the five colors (white, blue, red, brown and black) shown in Fig.\,\ref{copa}.
If the reflectance of the white paper is taken as 1.00, then the ratios of the reflectances for the other colors are 0.93 for blue, 0.92 for red, 0.86 for brown, and 0.57 for black.
Experiments are conducted under the Case\,1 and 2 conditions to compare the time to contact, the time difference of contact, and the behavior of the target position.
In addition, experiments are conducted not only when the surfaces of the objects have equal reflectance but also when they have different reflectances to investigate to what extent differences in reflectance can be allowed.
The above experiments are conducted for the case in which the object center position is $x_\text{m}=0.03\ \text{m}$.
%%In addition to the case where the reflectance of the object surface detected by the proximity sensor of each finger is equal, experiments are also conducted for the case where the reflectance is different in order to investigate the limitations of the proposed method.

\subsubsection{Experimental Results}
\begin{figure*}[htbp]
	\centering
	\begin{minipage}[b]{0.45\linewidth}
		\centering
		\includegraphics[width=76mm]{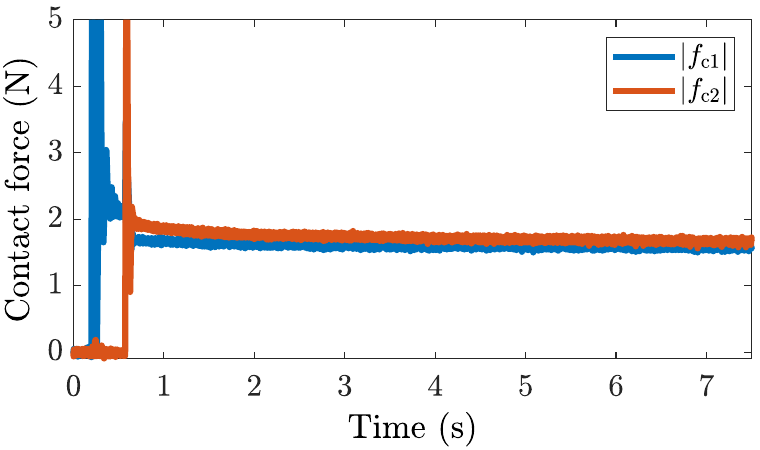}
		\subcaption{Time series data of the contact forces acting on each finger. The blue line shows the contact force acting on finger $j=1$, and the red line shows the contact force acting on finger $j=2$.\\  \ }
	\end{minipage}
         \hspace{0.04\columnwidth}
	\begin{minipage}[b]{0.45\linewidth}
		\centering
		\includegraphics[width=80mm]{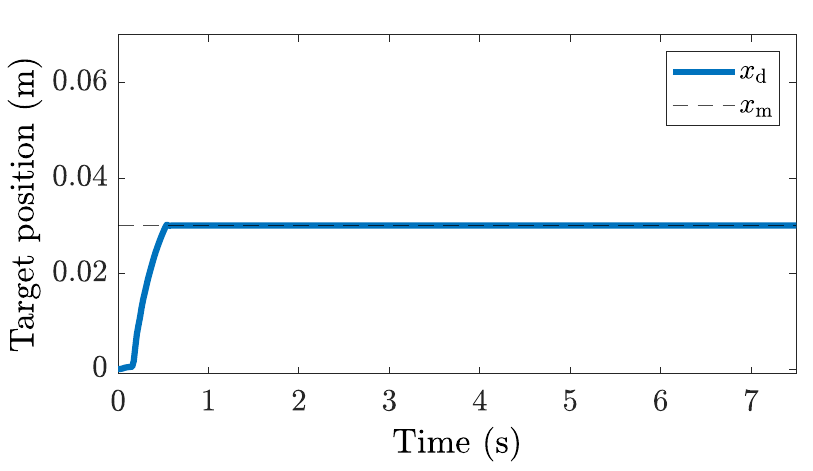}
		\subcaption{Time series data of the target position. The dashed line represents the center position of the grasped object, and the blue line represents the target position, i.e., the geometric center of gravity of the gripper.}
	\end{minipage}
	\caption{Experimental results without impact reduction and simultaneous contact control}
	\label{contact_result}
\end{figure*}
\begin{figure*}[htbp]
	\centering
	\begin{minipage}[b]{0.45\linewidth}
		\centering
		\includegraphics[width=77mm]{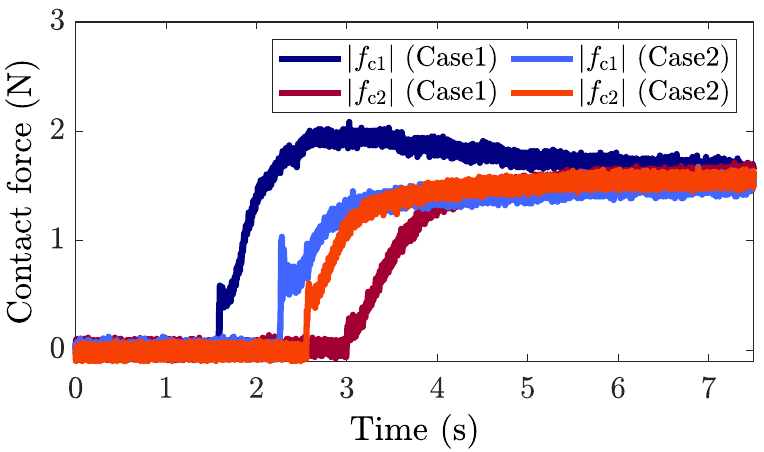}
		\subcaption{Time series data of the contact forces acting on each finger when the object surface is white in Case\,1 and Case\,2. The blue line shows the contact force acting on finger $j=1$, and the red line shows the contact force acting on finger $j=2$.\\ \ }
	\end{minipage}
         \hspace{0.04\columnwidth}
	\begin{minipage}[b]{0.45\linewidth}
		\centering
		\includegraphics[width=80mm]{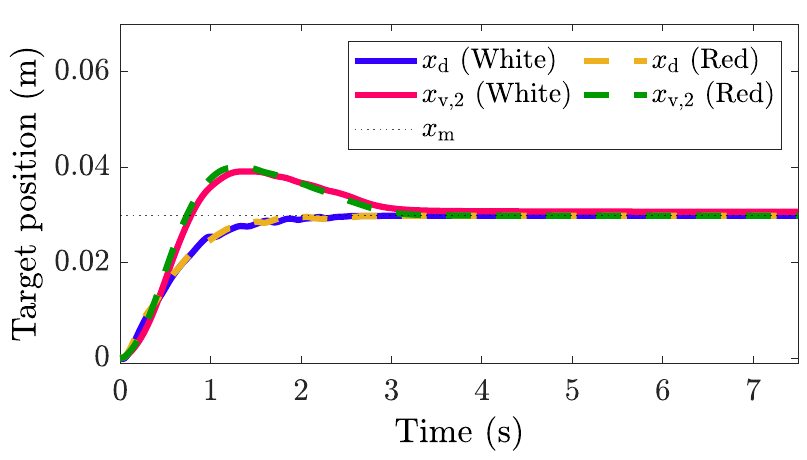}
		\subcaption{Time series data of the target position when the object surface is white and red in Case\,2. The dotted line represents the center position of the grasped object, the dashed line represents the results for the red surface, and the solid line represents the results for the white surface.}
	\end{minipage}
	\caption{Experimental results with and without simultaneous contact control.}
	\label{ex_comb}
\end{figure*}
The results for the contact force and target position in Case\,0 are shown in Fig.\,\ref{contact_result}, and the results in Case\,1 and Case\,2 are shown in Fig.\,\ref{ex_comb}.
Fig.\,\ref{contact_result}(a) shows that an impact force is generated in Case\,0, while it is not generated in Cases\,1 and 2 (Fig.\,\ref{ex_comb}(a)), in which impact reduction control is applied, indicating that the impact force is reduced.
\par
Next, we focus on the time difference of contact.
In addition to the results obtained in Fig.\,\ref{ex_comb}(a), the experimental results for $x_\text{m}=0.00, 0.01, 0.02, 0.04, 0.05\ \text{m}$ are also summarized in Table\,\ref{position_dep}.
Table\,\ref{position_dep} shows that the simultaneous contact function can be performed regardless of the object position.
\par
Then, we focus on the target position.
Fig.\,\ref{ex_comb}(b) shows that the target position $x_{\text{v},2}$ is overshot when adding simultaneous contact control.
This phenomenon is also confirmed in the simulation.
\par
Finally, we focus on the dependence on reflectance when the colors of the detected surfaces are the same.
A table summarizing the time to contact and the time difference of contact when the color of the object surface is varied is shown in Table\,\ref{color_dep}.
According to Table\,\ref{color_dep}, the time difference of contact can be reduced by adding simultaneous contact control.
In addition, the time to contact is almost independent of the color.
This can be explained by the fact that there is almost no change in the behavior of the target position $x_{\text{v},2}$ with the color of the object, as shown in Fig.\,\ref{ex_comb}(b).
Therefore, if the reflectances of the detected surfaces are equal, then the simultaneous contact function is independent of the reflectance.
The results for different colors of the detected surfaces are described in Sec.\,V.
\begin{table*}[htbp]
    \caption{Dependence of the simultaneous contact function on the grasped object position }
    \centering
    \scalebox{1.02}{
     \renewcommand{\arraystretch}{1.4}
    \begin{tabular}{cccccccc}
        \hline
        \multicolumn{1}{c|}{} & \multirow{3}{*}{$x_\text{m}$ (m)} &  \multicolumn{3}{|c}{Without simultaneous contact control  (Case\,1)} &  \multicolumn{3}{|c}{With simultaneous contact control  (Case\,2)} \\ \cline{3-8}
\multicolumn{1}{c|}{Evaluation method} &  & \multicolumn{2}{|c|}{Time to contact\ (s)} & \ Time difference\ & \multicolumn{2}{|c|}{Time to contact\ (s)} & \ Time difference\\ \cline{3-4} \cline{6-7}
\multicolumn{1}{c|}{} &  & \multicolumn{1}{|c|}{\ Finger\,1\ } &  \multicolumn{1}{c|}{\ Finger\,2\ } &\ of contact\ (s)\ & \multicolumn{1}{|c|}{\ Finger\,1\ } & \multicolumn{1}{c|}{\ Finger\,2\ } &\ of contact\ (s)\ \\  \hline\hline
\multicolumn{1}{c|}{} & \multicolumn{1}{c|}{\ 0.00\ } & \multicolumn{1}{|c|}{1.84} &  1.84 & \multicolumn{1}{|c|}{0.00} & 1.84 & \multicolumn{1}{|c|}{1.84} & 0.00 \\
\multicolumn{1}{c|}{} & \multicolumn{1}{c|}{\ 0.01\ } & \multicolumn{1}{|c|}{1.64} &  2.06 & \multicolumn{1}{|c|}{0.42} & 1.85 & \multicolumn{1}{|c|}{1.88} & 0.03 \\
\multicolumn{1}{c|}{Simulation} & \multicolumn{1}{c|}{\ 0.02\ } & \multicolumn{1}{|c|}{1.38} &  2.22 & \multicolumn{1}{|c|}{0.84} & 1.87 & \multicolumn{1}{|c|}{1.88} & 0.01 \\
\multicolumn{1}{c|}{} & \multicolumn{1}{c|}{\ 0.03\ } & \multicolumn{1}{|c|}{1.15} &  2.37 & \multicolumn{1}{|c|}{1.22} & 1.86 & \multicolumn{1}{|c|}{1.86} & 0.00 \\
\multicolumn{1}{c|}{} & \multicolumn{1}{c|}{\ 0.04\ } & \multicolumn{1}{|c|}{0.94} &  2.49 & \multicolumn{1}{|c|}{1.55} & 1.83 & \multicolumn{1}{|c|}{1.84} & 0.01 \\
\multicolumn{1}{c|}{} & \multicolumn{1}{c|}{\ 0.05\ } & \multicolumn{1}{|c|}{0.68} &  2.56 & \multicolumn{1}{|c|}{1.88} & 1.74 & \multicolumn{1}{|c|}{1.77} & 0.03 \\ \hline\hline
\multicolumn{1}{c|}{} & \multicolumn{1}{c|}{\ 0.00\ } & \multicolumn{1}{|c|}{2.32} &  2.58 & \multicolumn{1}{|c|}{0.26} & 2.14 & \multicolumn{1}{|c|}{2.31} & 0.17 \\
\multicolumn{1}{c|}{} & \multicolumn{1}{c|}{\ 0.01\ } & \multicolumn{1}{|c|}{1.81} &  3.02 & \multicolumn{1}{|c|}{1.21} & 2.21 & \multicolumn{1}{|c|}{2.35} & 0.14 \\
\multicolumn{1}{c|}{Experiment} & \multicolumn{1}{c|}{\ 0.02\ } & \multicolumn{1}{|c|}{1.73} &  3.18 & \multicolumn{1}{|c|}{1.45} & 2.27 & \multicolumn{1}{|c|}{2.47} & 0.20 \\
\multicolumn{1}{c|}{} & \multicolumn{1}{c|}{\ 0.03\ } & \multicolumn{1}{|c|}{1.59} &  3.00 & \multicolumn{1}{|c|}{1.41} & 2.26 & \multicolumn{1}{|c|}{2.55} & 0.29 \\
\multicolumn{1}{c|}{} & \multicolumn{1}{c|}{\ 0.04\ } & \multicolumn{1}{|c|}{1.45} &  4.29 & \multicolumn{1}{|c|}{2.84} & 1.92 & \multicolumn{1}{|c|}{2.30} & 0.38 \\
\multicolumn{1}{c|}{} & \multicolumn{1}{c|}{\ 0.05\ } & \multicolumn{1}{|c|}{1.34} &  4.81 & \multicolumn{1}{|c|}{3.47} & 1.81 & \multicolumn{1}{|c|}{2.20} & 0.39 \\ \hline
    \label{position_dep}
    \end{tabular}
 }
\end{table*}
\begin{table*}[htbp]
    \caption{Experimental results of the dependence of the simultaneous contact function on the grasped object reflectance}
    \centering
    \scalebox{1.02}{
     \renewcommand{\arraystretch}{1.4}
    \begin{tabular}{cccccccc}
        \hline
        \multicolumn{2}{c}{\multirow{2}{*}{Color}} &  \multicolumn{3}{|c}{Without simultaneous contact control  (Case\,1)} &  \multicolumn{3}{|c}{With simultaneous contact control  (Case\,2)} \\ \cline{3-8}
                              &     & \multicolumn{2}{|c|}{Time to contact\ (s)} & \ Time difference\ & \multicolumn{2}{|c|}{Time to contact\ (s)} & \ Time difference\\ \cline{1-4} \cline{6-7}
Finger\,1 side & \multicolumn{1}{|c}{Finger\,2 side} & \multicolumn{1}{|c|}{\ Finger\,1\ } &  \multicolumn{1}{c|}{\ Finger\,2\ } &\ of contact\ (s)\ & \multicolumn{1}{|c|}{\ Finger\,1\ } & \multicolumn{1}{c|}{\ Finger\,2\ } &\ of contact\ (s)\ \\  \hline\hline
White & \multicolumn{1}{|c}{White} & \multicolumn{1}{|c|}{1.59} & 3.00  & \multicolumn{1}{|c|}{1.41} & 2.26 & \multicolumn{1}{|c|}{2.55} & 0.29 \\
Red & \multicolumn{1}{|c}{Red} & \multicolumn{1}{|c|}{1.61} & 3.18  & \multicolumn{1}{|c|}{1.57} & 2.28 & \multicolumn{1}{|c|}{2.66} & 0.38 \\
Blue & \multicolumn{1}{|c}{Blue} & \multicolumn{1}{|c|}{1.61} & 3.49  & \multicolumn{1}{|c|}{1.88} & 2.35 & \multicolumn{1}{|c|}{2.74} & 0.39 \\
Brown & \multicolumn{1}{|c}{Brown} & \multicolumn{1}{|c|}{1.53} & 3.42  & \multicolumn{1}{|c|}{1.89} & 2.10 & \multicolumn{1}{|c|}{2.41} & 0.31 \\
Black & \multicolumn{1}{|c}{Black} & \multicolumn{1}{|c|}{1.55} & 3.32  & \multicolumn{1}{|c|}{1.77} & 2.01 & \multicolumn{1}{|c|}{2.40} & 0.39 \\ \hline
Brown & \multicolumn{1}{|c}{White} & \multicolumn{1}{|c|}{1.33} & 2.82  & \multicolumn{1}{|c|}{1.49} & 2.38 & \multicolumn{1}{|c|}{2.53} & 0.15 \\
White & \multicolumn{1}{|c}{Brown} & \multicolumn{1}{|c|}{1.40} & 2.92  & \multicolumn{1}{|c|}{1.52} & 2.62 & \multicolumn{1}{|c|}{2.43} & 0.19 \\ \hline
    \label{color_dep}
    \end{tabular}
 }
\end{table*}

\par
These results show that the proposed method is effective for realizing both simultaneous contact and impact reduction functions.
Although factors such as frictional forces and left-right differences in the reflectance even for the same color cannot be ignored in the experiment, the effectiveness of the proposed method is confirmed in the experiment, as well as in the simulation.

\section{Discussion of limitations}
A limitation of the proposed method is that the reflectances of the surfaces detected by the proximity sensor of each finger are assumed to be equal.
If the reflectances of the surfaces detected by the proximity sensor differ, then the target trajectory does not converge to the central position of the object, as can be seen from (\ref{fsa}).
However, the reflectances of the detected surfaces do not need to be exactly the same for the simultaneous contact function to perform.
Table \ref{color_dep} shows that simultaneous contact can be achieved even when the object surface colors are white and brown.
Moreover, when the difference in reflectance is smaller than that in this condition, the simultaneous contact function is almost equivalent to that for the same color of the detected surfaces.
In addition, there is almost no difference in the simultaneous contact function when the correspondence between the distance between each finger and the object and the magnitude of the reflectance are interchanged.
These results indicate that the simultaneous contact function is achieved even if there are slight differences in the pattern and color arrangement of the detected surfaces. (The proposed method is effective for grasping familiar objects.)
\par
In the simulations and experiments, the following conditions are imposed: the object surface is flat, the reflectance is constant over the entire plane, and the area of the object surface is sufficiently large compared to the detection range of the proximity sensor.
These conditions are imposed to facilitate evaluation and are not limitations of this method.
The proximity sensor used in this study is of the O-RLI type, so the intensity of the light received by the phototransistor corresponds to the sensor output.
In other words, the condition for the simultaneous contact function to perform is that the final received light intensities are equal to some extent.
If this condition is satisfied, then the proposed method can apply the simultaneous contact function by using only the received light intensities, which is the most primitive output of the optical method.

\section{CONCLUSION}
In this letter, we proposed a method for simultaneous contact of each finger, impact reduction, and contact force control of a gripper.
Simultaneous contact and impact reduction are achieved via precontact control using O-RLI-type proximity sensors.
Additionally, the multiple-impedance control framework enables the three functions to be achieved without switching control laws, and the addition of the new simultaneous contact function does not result in the loss of any other functions.
In the future, we intend to develop a method that does not depend on the difference in the reflectance of grasped surfaces and to apply the method to multifingered hands and soft hands.

\addtolength{\textheight}{-12cm}   % This command serves to balance the column lengths

%%%%%%%%%%%%%%%%%%%%%%%%%%%%%%%%%%%%%%%%%%%%%%%%%%%%%%%%%%%%%%%%%%%%%%%%%%%%%%%%

%%%%%%%%%%%%%%%%%%%%%%%%%%%%%%%%%%%%%%%%%%%%%%%%%%%%%%%%%%%%%%%%%%%%%%%%%%%%%%%%

%%%%%%%%%%%%%%%%%%%%%%%%%%%%%%%%%%%%%%%%%%%%%%%%%%%%%%%%%%%%%%%%%%%%%%%%%%%%%%%%

\end{document}